*Article*

# ASTEROID: A Spatiotemporal Information Transformer for Forecasting Multi-Step Time Series of Molecular Dynamics

Kexin Wu,[1] Luonan Chen,[2,*] and Renxiao Wang[1,*]

*1. Department of Medicinal Chemistry, School of Pharmaceutical Sciences, Fudan University, 826 Zhangheng Road, Shanghai 201203, People's Republic of China*

*2. School of Mathematical Sciences and School of AI, Shanghai Jiao Tong University, Shanghai 200240, People's Republic of China*

* To whom all correspondence should be addressed: *wangrx@fudan.edu.cn* (R. Wang); *lnchen@sjtu.edu.cn* (L. Chen)

## ABSTRACT

Molecular dynamics (MD) simulation is computationally demanding, particularly for large-scale systems requiring long-term analysis. Accurate forecast of the outcomes of a MD simulation is not only an attractive scientific challenge but also has substantial practical value. In this work, we developed a data-driven framework, termed ASTEROID (Advanced Spatiotemporal TransformER fOr Inferring Dynamics), that can directly predict multi-step atomic coordinates, avoiding conventional iterative integration. For this purpose, our ASTEROID reformulates MD trajectories as high-dimensional spatiotemporal sequences and integrates the Spatiotemporal Information (STI) Transformation equation into a Transformer architecture. The core innovation of ASTEROID lies in its ability to model multiscale spatiotemporal dependencies. In particular, for spatial dependencies, a local-global self-attention mechanism captures both short- and long-range interactions. For temporal dependencies, an encoder-decoder structure integrates global context with autoregressive forecasting. ASTEROID was evaluated on several quantum-mechanics derived molecular datasets. Our results indicate that ASTEROID achieved not only a higher level of accuracy in multi-step prediction than existing methods on various benchmarks, but also significantly reduced computational cost of conventional MD simulation. Moreover, the model supports iterative multi-step forecasting over an extended time scale. This work establishes a robust and generalizable data-driven paradigm for accelerating MD simulations.

## 1. INTRODUCTION

Molecular dynamics (MD) simulation has become an indispensable tool in biomacromolecules research[1]. It has enabled a paradigm shift from the study of static structures to the analysis of dynamic processes[2,3]. However, the limitations of molecular dynamics simulations arise primarily from the inherent complexity of the atomic system's potential energy surface and the high computational cost associated with calculating energy and forces across extended spatial and temporal scales[4,5,6]. Therefore, there is a compelling need to integrate artificial intelligence (AI) methodologies to accelerate these simulations[7].

Current AI-driven approaches in molecular dynamics simulation can be broadly classified into the following categories: (1) Quantum Mechanical Methods: These employ neural networks to parameterize wave functions or approximate density functionals within quantum mechanical frameworks[8-13]. (2) Parametric Models: This category encompasses machine learning force fields, coarse-grained models, and interatomic potentials[14-19]. (3) Time Series Forecasting Methods: These techniques leverage historical trajectory data to predict future states of the system, employing architectures such as recurrent or Transformer networks for temporal dynamics modeling[20-22]. (4) Enhanced Sampling Techniques: Methods such as Markov state models, clustering, and dimensionality reduction are utilized to integrate multiple short-time trajectories, thereby reconstructing long-time-scale dynamics and improving sampling efficiency[23-28]. (5) Conformation Generation Methods: These include diffusion models and other generative frameworks that directly produce plausible molecular structures[29-33].

Current computational methods in molecular dynamics face several inherent challenges. Parametric and quantum mechanical approaches exhibit notable limitations, including a heavy reliance on high-quality training data, limited generalization capability, and insufficient simulation stability[15-17]. Conformational exploration methods, on the other hand, are often constrained by a narrow predictive scope and the absence of unified validation frameworks[31-33]. In comparison, time-series forecasting offers a more direct modeling strategy. However, conventional time-series techniques often struggle to achieve sufficient temporal scale and predictive accuracy for molecular dynamics.

The principal challenges in time series forecasting stem from two fundamental issues. The first involves predicting subsequent time series segments rather than individual time points. Many prevailing methods are limited to short-term forecasting and struggle to extend the prediction horizon. Secondly, predictive performance tends to deteriorate significantly as the forecast horizon lengthens. Although deep learning models such as Recurrent Neural Networks (RNNs) and Long Short-Term Memory (LSTM) networks have advanced sequence modeling capabilities to some extent, they remain limited by issues like gradient vanishing and inadequate long-term memory retention, which impedes their application to long-term forecasting[34-36].

In recent years, Transformer models have offered a new solution for time series forecasting by effectively capturing long-range dependencies in historical data through self-attention mechanisms[37]. Subsequent improvements, encompassing both core mechanisms like attention optimization and architectural designs like hierarchical representations, have broadened the applicability of Transformer-based architectures in temporal forecasting tasks[38]. Nevertheless, these sophisticated models still face limitations when applied to the specific challenges of MD simulations. The inherent nonlinearity and high-dimensional nature of molecular dynamical systems pose significant challenges for accurate prediction. The task of multi-step forecasting becomes especially challenging under constraints of limited short-term observational data.

To address these constraints, we introduce the Spatiotemporal Information (STI) Transformation equation[39,40]. Based on the delay embedding theorem, the STI framework establishes a mapping from high-dimensional spatial information to future values of any target variable[41,42]. This mapping effectively broadens the scope of utilizable data, thereby alleviating the curse of dimensionality caused by limited sample sizes. However, due to the inherent high-dimensionality and strong nonlinearity of the equation, current numerical methods face significant challenges in achieving the required levels of accuracy and robustness for practical forecasting applications..

This study presents a novel methodology that integrates the STI equation with time-series forecasting to accelerate molecular dynamics simulations. Our approach directly predicts atomic coordinates multiple steps ahead, thereby circumventing the computational bottleneck of numerical integration inherent in traditional classical molecular dynamics and quantum mechanical (QM) simulations. The proposed ASTEROID framework leverages the STI transformation to map high-dimensional spatiotemporal states onto the future values of target variables. The model employs an encoder-decoder architecture to process high-dimensional spatiotemporal data, integrating both spatial and temporal information. This achieves two primary objectives: the effective expansion of the dataset for target variables and the autonomous generation of multi-step future predictions.

To comprehensively assess the performance of ASTEROID, we evaluated it against multiple real-world datasets. The results demonstrate that ASTEROID outperforms various baseline and deep learning methods in multi-step forecasting, offering better accuracy and robustness. In practice, ASTEROID provides accurate predictions across diverse datasets and achieves a substantial reduction in the computational cost required to generate quantum-accurate MD trajectories. This framework thus presents a novel and efficient approach for accelerating molecular dynamics simulations.

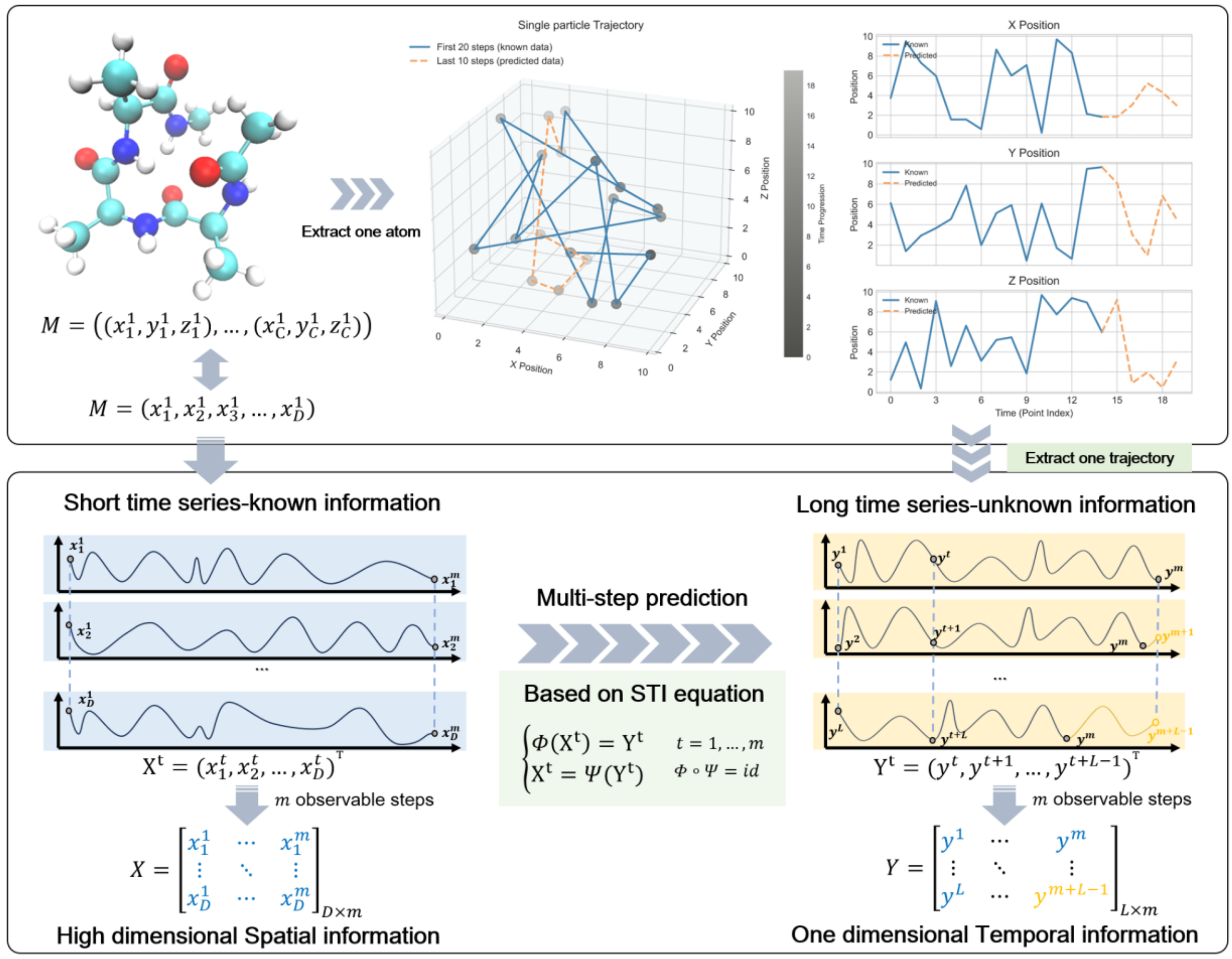

**Figure 1.** Schematic diagram of the ASTEROID model. The ASTEROID framework models and accelerates molecular dynamics by representing the XYZ coordinates of each atom as a high-dimensional spatiotemporal series $\mathbf{X}^t$ and employing time series forecasting. The target variable $y^t$ corresponds to a specific coordinate time series of a selected atom. Predicting its future state from short-term, high-dimensional observations remains challenging. To capture the temporal dependencies inherent in the target variable $y^t$, we apply the delay-embedding strategy. Applying the delay embedding strategy, we establish the correspondence between the observational vector $\mathbf{X}^t = (x_1^t, x_2^t, \ldots, x_D^t)^T$; and the delay coordinate vector $\mathbf{Y}^t=(y^t, y^{t+1}, \ldots, y^{t+L-1})^T$. This relationship forms the Spatiotemporal Information (STI) Transformation equation, which comprises both primal and conjugate forms. When the observable series contains m time steps, both $X$ and $Y$ can be expressed as matrices. Thus, given historical data $X$, an L−1-step ahead prediction of $y^t$ is achieved by solving for $\Phi$ and $Y$ within the STI framework. To handle the inherent nonlinearity and high dimensionality of molecular systems, ASTEROID integrates the STI equation with a Transformer-based spatiotemporal attention architecture, resulting in a unified inference framework.

## 2. METHODS

### 2.1 Delay embedding theorem

Consider a nonlinear dynamical system whose state evolves according to

$$\mathbf{X}^{t+1} = \phi(\mathbf{X}^{t})$$

where $\phi$ is a nonlinear map, $\mathbf{X}^t = (x_1^t, x_2^t, \ldots, x_D^t)^T$ denotes the D-dimensional state vector at time t (here refers to atomic coordinates), and the superscript T indicates transpose. The indices t and t+1 denote consecutive, discrete time points. After prolonged evolution, the system converge to stable or recurrent state known as an attractor.

The mathematical basis of phase space reconstruction is the time-delay embedding theorem and its generalizations[41,42]. This theorem posits that the evolution of any single variable in a multi-variable dynamical system is governed by its coupling with other variables. Consequently, a time series of a single variable is sufficient to reconstruct a phase space that is topologically equivalent to that of the original system.

Specifically, Takens's theorem guarantees that, whenever the embedding dimension satisfies L>2d>0, where d is the box-counting dimension of the system's attractor, the delay embedding vector $\mathbf{Y}^t = (y^t, y^{t+1}, \ldots, y^{t+L-1})^T$ can reconstruct the dynamics of the original high-dimensional system $\mathbf{X}^t$.

$x_1^t$in a topologically equivalent manner. Here, $\mathbf{X}^t$ has D variables, and L is the embedding dimension of $\mathbf{Y}^t$.

$$\Phi(\mathbf{X}^t) = \mathbf{Y}^t = (y^t, y^{t+1}, \ldots, y^{t+L-1})^T$$

Takens's theorem provides a mathematical basis for reconstructing the phase space of a high-dimensional dynamical system from a scalar time series. The theorem strictly requires the original system to possess a compact, smooth attractor. However, this condition is frequently violated in real-world complex systems, especially those characterized by high dimensionality, nonlinearity, and inherent stochasticity. Nevertheless, in practice, systems such as molecular dynamics, climate models, and neural activity often show exhibit statistically stable over time. They tend to settle into a bounded, low-dimensional region. Although this bounded region does not constitute a perfect mathematical attractor, it nevertheless fosters stable, long-term statistical regularity, which enables feasible prediction.

Consequently, even when the strict theoretical prerequisites of Takens's theorem are not fully met, delay embedding methods find broad application in the analysis and forecasting of real-world systems. The approach reconstructs a topologically equivalent phase space using only a one-dimensional observational signal. This reconstructed space preserves the essential dynamical invariants of the original system, thereby facilitating accurate predictions. The efficacy of this data-driven methodology has been empirically demonstrated across diverse fields, including climate science, traffic flow analysis, and genomic studies[44-46].

## 2.2 The STI equation

The STI equation is formally called the Spatiotemporal Information Transformation equation, a set of equations grounded in nonlinear dynamics. It establishes a mapping from the high-dimensional spatial information of a system to the future temporal evolution of any target variable. This transformation is particularly useful for understanding and predicting the dynamic behavior of complex systems when only short-term high-dimensional data are available.

The theory begins by considering two variables, $\mathbf{X}^t$ and $\mathbf{Y}^t$. Here, $\mathbf{X}^t$ represents high-dimensional multivariate spatial data, while $\mathbf{Y}^t$ denotes univariate time-series data observed at multiple consecutive time points from t to t+L-1. This single variable corresponds to one of the D variables in the system, and L denotes the embedding dimension. Given an observed time series comprising m time steps, both $X$ and $Y$ can be represented in matrix form.

$$X = [\mathbf{X}^1, \mathbf{X}^2, \dots, \mathbf{X}^m] = \begin{bmatrix} x_1^1 & \cdots & x_1^m \\ \vdots & \ddots & \vdots \\ x_D^1 & \cdots & x_D^m \end{bmatrix}_{D\times m}$$

$$Y = \begin{bmatrix} y^1 & \cdots & y^m \\ \vdots & \ddots & \vdots \\ y^L & \cdots & y^{m+L-1} \end{bmatrix}_{L\times m}$$

Based on the previous derivation, we obtain $\Phi(X) = Y$, where $\Phi = [\Phi_1, \Phi_2, \dots, \Phi_L]$. Since embedding is a one-to-one mapping, the high-dimensional attractor and the reconstructed delay attractor with appropriately chosen dimensions are topologically conjugate. This conjugacy is mathematically expressed as follows:

$$X = \Phi^{-1}(Y) = \Psi(Y)$$

Thus, we construct the following STI transformation equation. This equation maps D-dimensional data $\mathbf{X}^t$ to L-dimensional data $\mathbf{Y}^t$, enabling bidirectional conversion between high-dimensional spatial information and low-dimensional temporal information:

$$\begin{cases} \Phi(X) = Y \\ X = \Psi(Y) \end{cases} \quad t = 1, \dots, m$$

Here, $\Phi$ and $\Psi$ are conjugate function. Therefore, we can obtain the following:

$$\begin{bmatrix} \Phi_1(x_1^1) & \Phi_1(x_1^2) & \cdots & \Phi_1(x_1^m) \\ \Phi_2(x_2^1) & \Phi_2(x_2^2) & \cdots & \Phi_2(x_2^m) \\ \vdots & \vdots & \ddots & \vdots \\ \Phi_L(x_D^1) & \Phi_L(x_D^2) & \cdots & \Phi_L(x_D^m) \end{bmatrix} = \begin{bmatrix} y^1 & y^2 & \cdots & y^m \\ y^2 & y^3 & \cdots & \hat{y}^{m+1} \\ \vdots & \vdots & \ddots & \vdots \\ y^L & y^{L+1} & \cdots & \hat{y}^{m+L-1} \end{bmatrix}$$

Given observed variables up to time step m, where "∧" denotes predicted values, the task requires learning the mappings $\Phi_i$ (where i = 1, 2, …, L) from available $\mathbf{X}^t$ (t=1, …, m) while simultaneously predicting target variable y from steps M+1 to M+L-1. Thus, given $\mathbf{X}^t$, solving for $\Phi_i$ and $\mathbf{Y}^t$ in the formula enables an (L-1)-step-ahead prediction of $y$.

While the original system may be high-dimensional (with dimension D), the intrinsic dimension d of its steady state or underlying manifold is typically much lower in practice, such that D >> d. Consequently, a relatively low embedding dimension L is often sufficient to capture the system's essential dynamics. Computationally, it is standard practice to choose L such that it exceeds twice the intrinsic dimension, that is, L > 2d.

Based on a deep learning framework, both the spatial embedding mapping Φ and the temporal prediction mapping Ψ can be learned. This enables the model to predict long-term, low-dimensional temporal behavior from short-term, high-dimensional spatial data, thereby facilitating effective multi-step forecasting. Current implementations integrating the STI equation for time-series forecasting in nonlinear dynamical systems include models such as the auto-reservoir neural network (ARNN), the spatiotemporal information conversion machine (STICM), and delay-embedding-based forecast machine (DEFM)[44-46]. Nevertheless, these models have primarily been applied to idealized systems (e.g., the Lorenz system) and have not yet been successfully implemented in complex, real-world systems such as molecular dynamics simulations.

### 2.3 Data preparation and embedding

This study employed the MD22 and MD_analysis datasets[47,48]. These trajectories were denoised with a Savitzky-Golay filter and preprocessed through coordinate separation and min-max normalization. The resulting time-series data was formatted for spatiotemporal neural network training.

The model incorporated embedding layer comprising three primary components: TokenEmbedding, PositionalEncoding, and ElementEmbedding. These components worked collectively to map the raw input sequence into a high-dimensional representation, effectively capturing structural, positional, and elemental information.

Specifically, the TokenEmbedding component transformed the original input $X \in \mathbb{R}^{T \times N}$ into an extended tensor $TE \in \mathbb{R}^{T \times N \times d}$ via a linear transformation layer. The weight matrix of this layer was initialized using the Xavier uniform initialization method[50].

$$\mathbf{TE} = \mathrm{Linear}(\mathbf{X})$$

Here, Linear denoted a bias-free linear transformation layer.

The PositionalEncoding module implemented the sinusoidal positional encoding, generating unique positional identifiers $PE \in \mathbb{R}^{T \times 1 \times d}$ for each time step in the sequence[37]. This encoding was computed as follows:

$$\mathbf{PE}(pos, 2i) = \sin\left(\frac{pos}{10000^{2i/d}}\right), \mathbf{PE}(pos, 2i+1) = \cos\left(\frac{pos}{10000^{2i/d}}\right)$$

To further incorporate semantic information such as element type or amino acid species, we designed the ElementEmbedding module. This module took one-hot encodings e of atomic types as input and maps them to dense vectors through a trainable embedding matrix $W_{emb} \in \mathbb{R}^{N_e \times d}$, where $N_e$ denoted the total number of element types and d represented the embedding dimension:

$$\mathbf{EE} = \boldsymbol{W_{emb}}[\boldsymbol{e}]$$

We summed the outputs of the three components and applied dropout. This resulted in the unified embedded representation $X_{emb}$, which integrates token, positional, and elemental information. The representation served as the input to subsequent encoder layers.

$$\mathbf{X}_{emb} = \mathrm{Dropout}(\mathbf{TE} + \mathbf{PE} + \mathbf{EE})$$

### 2.4 The ASTEROID model

The Transformer architecture has demonstrated formidable capabilities in sequence modeling for time-series forecasting. By utilizing self-attention mechanisms, the model dynamically captured dependencies between any two time points in the time series without relying on the recursive structure of traditional recurrent neural networks (RNNs). Simultaneously, its globally connected structure overcomes the locality constraints of convolutional neural networks (CNNs). These capabilities translate into two core advantages: efficient parallelized computation and a superior ability to model long-range dependencies. The multi-head attention mechanism facilitates the simultaneous capture of patterns at different temporal scales, while positional encoding preserves the sequential order of time steps[37]. In time series forecasting, the Transformer typically adopted an encoder-decoder architecture. The encoder mapped historical sequences into latent representations, and the decoder performed multi-step predictions based on these representations and known outputs. To adapt to the characteristics of time series data, researchers often have introduced modifications such as seasonal-trend decomposition, sparse attention mechanisms to reduce computational complexity, and explicit temporal feature encoding to incorporate domain-specific temporal knowledge[51-53].

In our ASTEROID model, based on the principles of the STI equation, the encoder and decoder of the Transformer were utilized. The encoder input $\mathbf{X}_{emb}^{t}$ consisted of high-dimensional multivariate spatial vectors. This represented the coordinates of multiple atoms at the same time step t to extract effective spatial information. This information is then passed to the decoder. The decoder input $\overline{\mathbf{Y}}^{t}$ started with zero values and included the target variable's atomic coordinates at multiple time steps [t, t+1, …, t+L-2]. It captured the temporal evolution information of the target variable, expressed as $[0, y^t, y^{t+1}, \ldots, y^{t+L-2}]$. The decoder integrated the spatial information from the input variable $\mathbf{X}_{emb}^{t}$ with the temporal information from the target variable sequence $\overline{\mathbf{Y}}^{t}$. It thus produced a single-step forecast, $\hat{\boldsymbol{y}}^{t+L-1}$, for the target variable at the immediate future time step t+L-1. Multi-step predictions were then generated using an autoregressive inference mechanism based on a dynamic sliding window approach.

$$\begin{bmatrix} \Phi_1(x_1^1) & \Phi_1(x_1^2) & & \Phi_1(x_1^m) \\ \Phi_2(x_2^1) & \Phi_2(x_2^2) & \vdots & \Phi_2(x_2^m) \\ \cdots & \cdots & & \cdots \\ \Phi_L(x_D^1) & \Phi_L(x_D^2) & & \Phi_L(x_D^m) \end{bmatrix} = \begin{bmatrix} y^1 & y^2 & & y^m \\ y^2 & y^3 & \vdots & \hat{y}^{m+1} \\ \cdots & \cdots & & \cdots \\ y^L & y^{L+1} & & \hat{y}^{m+L-1} \end{bmatrix}$$

The model incorporated a sophisticated spatiotemporal attention mechanism consisting of three core components: temporal attention, short-range spatial attention, and long-range spatial attention. Since traditional Transformer models compute attention purely through data-driven methods, it might learn spurious correlations that violate physical laws. Our design incorporates physical constraints through two specialized masks: a short-range mask based on interatomic distances and a long-range mask derived from Dynamic Time Warping (DTW) similarity[54]. This differentiated architecture enabled comprehensive and efficient capture of multi-scale, hierarchical interactions within molecular systems.

In the spatial dimension, the model captured both local and global structural features through a multi-scale attention mechanism. Short-range attention modeled local physical interactions among adjacent atoms, while long-range attention identified dynamically correlated atom pairs despite significant spatial separation. In the temporal dimension, the model combined the encoder's global context modeling with the decoder's causal constraints, thereby ensuring coherent representation learning and prediction. The modular architecture provided strong scalability and flexible configuration, enabling straightforward adaptation to molecular systems of varying scales and complexities.

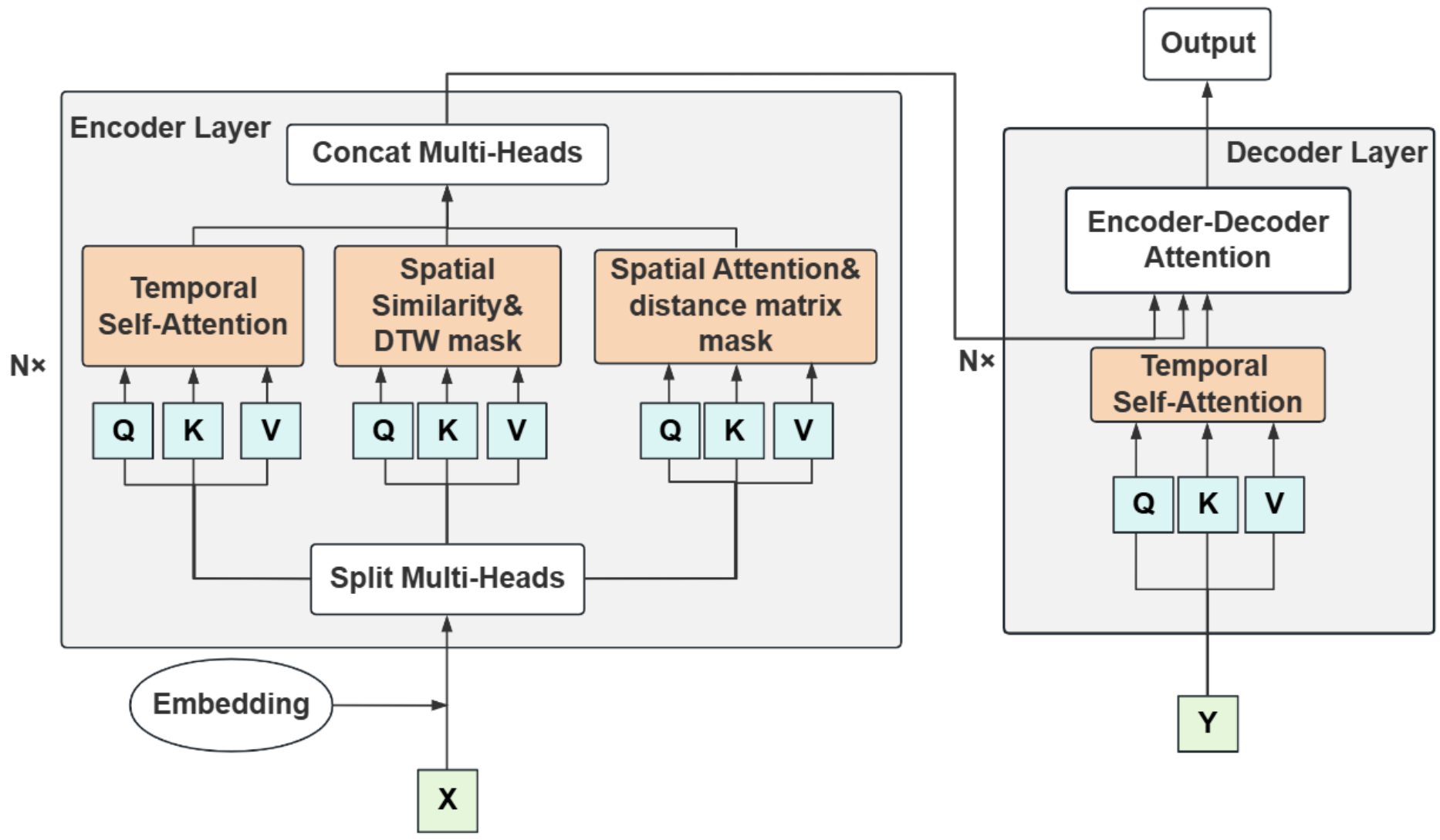

**Figure 2**. Architecture of the encoder-decoder model with the spatiotemporal attention mechanism. The model employs a differentiated attention scheme comprising temporal, short-range spatial, and long-range spatial components, integrated with masks based on physical distance and Dynamic Time Warping (DTW) similarity.

**Encoder**

In the encoder section, the model integrated three types of self-attention mechanisms in parallel: temporal attention, short-range spatial attention, and long-range spatial attention. For spatial modeling, a dual-scale attention mechanism was adopted to capture interactions at different ranges within molecular systems. Short-range spatial attention constructed an attention mask based on the Euclidean distance between atoms, focusing on modeling interactions among physically proximate atoms. Long-range spatial attention employed Dynamic Time Warping (DTW) as a similarity metric to identify correlations between spatially distant trajectories sharing similar dynamic characteristics. For temporal modeling, the encoder's temporal attention leveraged full connectivity across time steps to capture long-range dependencies within sequences.

In the specific implementation of the multi-head attention mechanism, a dedicated proportion of feature dimensions was allocated to each type of attention. The dimensionality of each head was determined by dividing the embedding dimension by the corresponding number of heads, with the square root of this dimension serving as a scaling factor to modulate attention weights. For each attention type, the model used independent two-dimensional convolutional layers to generate the query (Q), key (K), and value (V) matrices. The number of output channels for these layers was determined by both the number of heads $N_h$ and the embedding dimension $d_{model}$.

For each attention head h, its dimension $d_h$ was calculated as follows:

$$d_h = \frac{d_{model}}{N_h}$$

In the processing of spatial attention, the attention computation was performed along the atomic dimension. For each time step t, the model calculated attention scores between all atom pairs. The short-range spatial attention first computed a Euclidean distance matrix between atoms. For each time step, the model calculated the physical distances between all atom pairs and generates a binary mask matrix based on a predefined maximum distance threshold. Initially, the mask matrix was configured to reference the distance thresholds at each individual time step. However, it was observed that the distance matrices across different time steps exhibited minimal variation. The distance matrix from the initial time step was subsequently adopted to generate the mask. This mask matrix was used to filter out attention computations between atom pairs whose distances exceed the threshold. The study revealed that in smaller molecular systems, a strict masking approach could lead to computational errors when entire rows of the mask matrix were assigned "true" values. Conversely, an overly small threshold risked excluding crucial interactions. Therefore, for small-scale datasets, disabling this masking mechanism was recommended.

$$D(t, i, j) = \sqrt{\sum_{k=1}^{3} (x_k(t, i) - x_k(t, j))^2}$$

Here, t denoted a specific time step, and i, j represented different atoms, where x signified the spatial coordinates of an atom. A binary mask $\boldsymbol{M}_{\text{short}}$ was created by applying masking if the distance D(t, i, j) exceeded a predefined maximum.

$$\boldsymbol{Q}_{short} = \text{Conv2D}(\mathbf{X}), \qquad \boldsymbol{K}_{short} = \text{Conv2D}(\mathbf{X}), \qquad \boldsymbol{V}_{short} = \text{Conv2D}(\mathbf{X})$$

$$\boldsymbol{A}_{short} = \text{Softmax}\left(\frac{\boldsymbol{Q}_{short}\boldsymbol{K}_{short}^{T} + \boldsymbol{M}_{short}}{\sqrt{d_k}}\right)\boldsymbol{V}_{short}$$

Long-range spatial attention employed the Dynamic Time Warping (DTW) algorithm to compute pairwise similarities between time series and construct a DTW distance matrix $\boldsymbol{M}_{\text{long}}$. The DTW algorithm quantified similarity through dynamic temporal alignment, effectively capturing inherent patterns in time series data. Based on the resulting DTW distance matrix, the most similar sequence pairs were selected according to a predefined similarity threshold.

The specific implementation of the DTW algorithm was as follows:(1) Initialize a distance matrix by iterating over each point of two time series and calculating the Euclidean distance between every pair of elements. (2) Starting from the bottom-right corner of this matrix, the algorithm backtracked toward the top-left corner, at each step selecting the path with minimal cumulative cost. (3) The final DTW distance was defined as the minimal cumulative cost along the optimal warping path[54]:

$$\mathrm{DTW}(i,j) = \min_{\pi}\left(\sum_{(t,t')\in\pi} \mathrm{dist}\left(\mathrm{trajectory}_i(t), \mathrm{trajectory}_j(t')\right)\right)$$

where $\pi$ represented the optimal alignment path.

$$\boldsymbol{Q}_{long} = \mathrm{Conv2D}(\mathbf{X}), \qquad \boldsymbol{K}_{long} = \mathrm{Conv2D}(\mathbf{X}), \qquad \boldsymbol{V}_{long} = \mathrm{Conv2D}(\mathbf{X})$$

$$\boldsymbol{A}_{long} = \mathrm{Softmax}\left(\frac{\boldsymbol{Q}_{long}\boldsymbol{K}_{long}^{T} + \boldsymbol{M}_{long}}{\sqrt{d_k}}\right)\boldsymbol{V}_{long}$$

Temporal attention was designed to model the evolutionary dynamics of individual atomic trajectories. In contrast to spatial attention, it operated along the temporal dimension, computing attention scores between different time steps for each trajectory. Within the encoder layers, temporal attention employed bidirectional connectivity, allowing information exchange across all time steps in the sequence.

$$\boldsymbol{Q}_{t1} = \mathrm{Conv2D}(\mathbf{X}), \qquad \boldsymbol{K}_{t1} = \mathrm{Conv2D}(\mathbf{X}), \qquad \boldsymbol{V}_{t1} = \mathrm{Conv2D}(\mathbf{X})$$

$$\boldsymbol{A}_{t1} = \mathrm{Softmax}\left(\frac{\boldsymbol{Q}_{t1}\boldsymbol{K}_{t1}^{T}}{\sqrt{d_k}}\right)\boldsymbol{V}_{t1}$$

Finally, the outputs from the three attention mechanisms were concatenated and linearly projected to the target dimension, achieving integrated feature representation.

$$\mathbf{X}_{enc} = \mathrm{LayerNorm}\left(\mathrm{Linear}\left(\mathrm{Concat}\left[\boldsymbol{A}_{short}, \boldsymbol{A}_{long}, \boldsymbol{A}_{t1}\right]\right)\right)$$

**Decoder**

To ensure the causality of the generation process and prevent future information leakage, the model implemented a triple safeguard at both the structural and inference mechanism levels.

Structurally, the decoder employed masked self-attention, which applied an upper-triangular mask matrix $\boldsymbol{M}_{\mathrm{causal}}$ to ensure that each position in the sequence could only attend to previous positions.

At the input sequence level, the decoder utilized a zero-prefixed sequence design. For target sequence generation of length L during inference, the input was constructed as $[0, y^{t}, y^{t+1}, \ldots, y^{t+L-2}]$, where the leading zero served as a placeholder for the initial time step.

$$\mathbf{Q}_{t2} = \mathrm{Conv2D}(\mathbf{X}), \qquad \mathbf{K}_{t2} = \mathrm{Conv2D}(\mathbf{X}), \qquad \mathbf{V}_{t2} = \mathrm{Conv2D}(\mathbf{X})$$

$$\boldsymbol{A}_{t2} = \text{Softmax}\left(\frac{\mathbf{Q}_{t2}\mathbf{K}_{t2}^{T} + \mathbf{M}_{causal}}{\sqrt{d_k}}\right)\mathbf{V}_{t2}$$

$$\boldsymbol{X}_{dec} = \text{LayerNorm}(\boldsymbol{X} + \boldsymbol{A}_{t2})$$

Subsequently, the output of the decoder $\mathbf{X}_{dec}$ served as the query, while the output of the encoder $\mathbf{X}_{dec}$ acted as the key and value, enabling cross-attention to leverage the contextual information from the encoder.

$$\boldsymbol{Q}_{enc-dec} = \text{Conv2D}(\mathbf{X}_{dec}) \quad \boldsymbol{K}_{enc-dec} = \text{Conv2D}(\mathbf{X}_{enc}) \quad \boldsymbol{V}_{enc-dec} = \text{Conv2D}(\mathbf{X}_{enc})$$

$$\boldsymbol{A}_{enc-dec} = \text{Softmax}\left(\frac{\boldsymbol{Q}_{enc-dec}\boldsymbol{K}_{enc-dec}^{T}}{\sqrt{d_k}}\right)\boldsymbol{V}_{enc-dec}$$

During inference, the model employed an autoregressive generation scheme with a dynamic sliding window. This mechanism ensured that each prediction step depended solely on historical observations and the model's previous outputs.

Specifically, the decoder was initialized with a segment of ground-truth historical sequences. At each iteration, the model predicted the subsequent value using the current window, appended this prediction to the window, and removed the earliest time step to maintain a fixed window length. This iterative process can be formally expressed as::

$$\mathrm{y}_{tmp}^{(n)} = \text{Concat}\left(y_{tmp}^{(n-1)}[1:], \hat{y}^{(n)}\right)$$

Where $y_{tmp}^{(n-1)}$ represented the input window from the previous time step, and $\hat{y}^{(n)}$ denoted the current predicted value.

By integrating the masked attention mechanism with a dynamic autoregressive inference strategy, the model guaranteed that all forecasts strictly obeyed the principle of temporal causality.

### 2.5 Model training and evaluation

#### Loss function

During model training, we employed a composite loss function ($L$) consisting of two key components: a primary Root Mean Square Error (RMSE) term and a smoothing regularization term:

$$\mathcal{L} = \mathcal{L}_{RMSE} + \lambda \cdot \mathcal{L}_{smooth}$$

Here, the hyperparameter λ balanced the relative importance of the two loss terms, ensuring the model maintains prediction accuracy without excessively compromising trajectory smoothness. $L_{RMSE}$ represented the RMSE between the predicted output and the ground truth, calculated as:

$$\mathcal{L}_{\mathrm{RMSE}} = \frac{1}{N}\sum_{i=1}^{N} (y_i - \hat{y}_i)^2$$

This term ensured the model's predictions remain highly consistent with the actual trajectory data and served as the primary optimization objective.

The second term, $L_{smooth}$, was a smoothing regularization term implemented by computing the mean absolute difference between adjacent time steps in the predicted sequence:

$$\mathcal{L}_{\mathrm{smooth}} = \frac{1}{T-1}\sum_{t=1}^{T-1} |\hat{y}_{t+1} - \hat{y}_t|$$

Here, T was the sequence length. This regularization was grounded in the physical prior of smooth atomic motion. Its function was to penalize kinetically implausible and abrupt changes in the ensuing predictions. The result was an enhancement in the physical plausibility of the generated trajectories.

**Model training**

The model was optimized using a three-phase protocol. Initially, a hold-out validation segment was reserved from the training data to calibrate essential training hyperparameters, ensuring stable convergence. Subsequently, we performed architectural hyperparameter optimization to identify the optimal model configuration. Finally, the model was retrained on the complete training dataset using the selected hyperparameters. The resulting model's generalization capability was assessed on a completely held-out test set that had no involvement in either training or hyperparameter selection.

**Evaluation Metrics**

The evaluation employed three established time-series forecasting metrics: Pearson correlation coefficient, RMSE, and NRMSE. The Pearson correlation quantified the linear association between predictions and ground truth, with values approaching 1 indicating stronger agreement. RMSE measured the absolute magnitude of prediction errors, where lower values denote higher accuracy. NRMSE normalized the RMSE by the standard deviation of the true values, providing a scale-independent measure of relative error, with lower values reflecting better precision.

**Baseline models**

The proposed model was evaluated against three categories of baseline methods: classical approaches (SVR, RNN, LSTM, HES[35,55-57]), STI equation-based models (STNN, STICM[45,58]), and contemporary deep learning architectures (DLinear, PatchTST[59,60]). All models were assessed using identical data splits, with detailed descriptions available in Supplementary Information.

## 3. RESULTS AND DISCUSSION

### 3.1 Data processing

To validate the effectiveness of the designed model, ASTEROID was first applied to four distinct spatiotemporal chaotic systems. These included three systems derived from the quantum mechanics-accurate MD22 dataset, namely tetrapeptide, DHA and Stachyose, as well as a 19-peptide system from the MD_analysis dataset. These four datasets encompassed data at both quantum mechanics (QM) and molecular mechanics (MM) accuracy levels and covered three types of biomolecules, specifically proteins, lipids, and carbohydrates.

The data were sampled at temperatures between 400 K and 500 K with a resolution of 1 fs, and the corresponding potential energies and atomic forces were computed at the PBE+MBD theory level. The MD_analysis dataset featured a 19-peptide sequence (AAAQAAQAQWAQRQATWQA) with an α-helix secondary structure, sourced from the molecular dynamics open-source analysis software MDAnalysis. For the MD22 dataset, where trajectories were recorded at 1 fs intervals, hydrogen atoms were removed, and all atomic trajectories were sampled at 500 fs intervals. For the MD_analysis dataset, where records were taken every 240 fs, C-α atoms of amino acids were sampled at 480 fs intervals.

The significant noise inherent in raw molecular dynamics simulation data poses considerable challenges for direct time-series forecasting. Therefore, data denoising is necessary. Savitzky-Golay(S-G) filter achieved denoising by performing least-squares polynomial fitting within a local window. It could effectively reduce noise without altering the overall shape and characteristics of the data. The S-G filter involved two key parameters: window size and polynomial order. The window size determined the number of data points used in the fitting process. Larger windows provide better smoothing but might blur data features. The polynomial order dictated the complexity of the fitting polynomial. Higher orders could better capture local variations but might lead to overfitting if excessively high.

For the tetrapeptide data, a window size of 5 and a polynomial order of 3 resulted in acceptable data error levels and effective filtering, enabling the model to make accurate predictions. Under these parameters, the performance metrics were as follows: MSE = 0.04695, SNR (Signal-to-Noise Ratio) = 19.07 dB, and Correlation = 0.9929.

### 3.2 Model performance

## Prediction results

Table 1 summarizes the model's forecasting performance across the four molecular systems. The reported values represent the aggregate performance, calculated as the mean values across all atomic trajectories within each system.

**Table 1.** The details of the datasets and prediction results of the model across four different biomolecular systems.

| Dataset | Formula and chemical structure | Trajectory | Embedding dimension | Pearson *R* | RMSE | NRMSE | Trajectory |
|---|---|---|---|---|---|---|---|
| Tetrapeptide | $C_{12}H_{22}N_4O_4$ | 60 | 10 | 0.706 | 1.063 | 1.132 | 60 |
| DHA | $C_{22}H_{32}O_2$ | 72 | 6 | 0.445 | 0.645 | 2.906 | 72 |
| Stachyose | $C_{24}H_{42}O_{21}$ | 135 | 8 | 0.361 | 0.694 | 1.870 | 135 |
| 19-peptide | AAAQAAQAQW–AQRQATWQA | 57 | 7 | 0.513 | 0.626 | 3.672 | 57 |

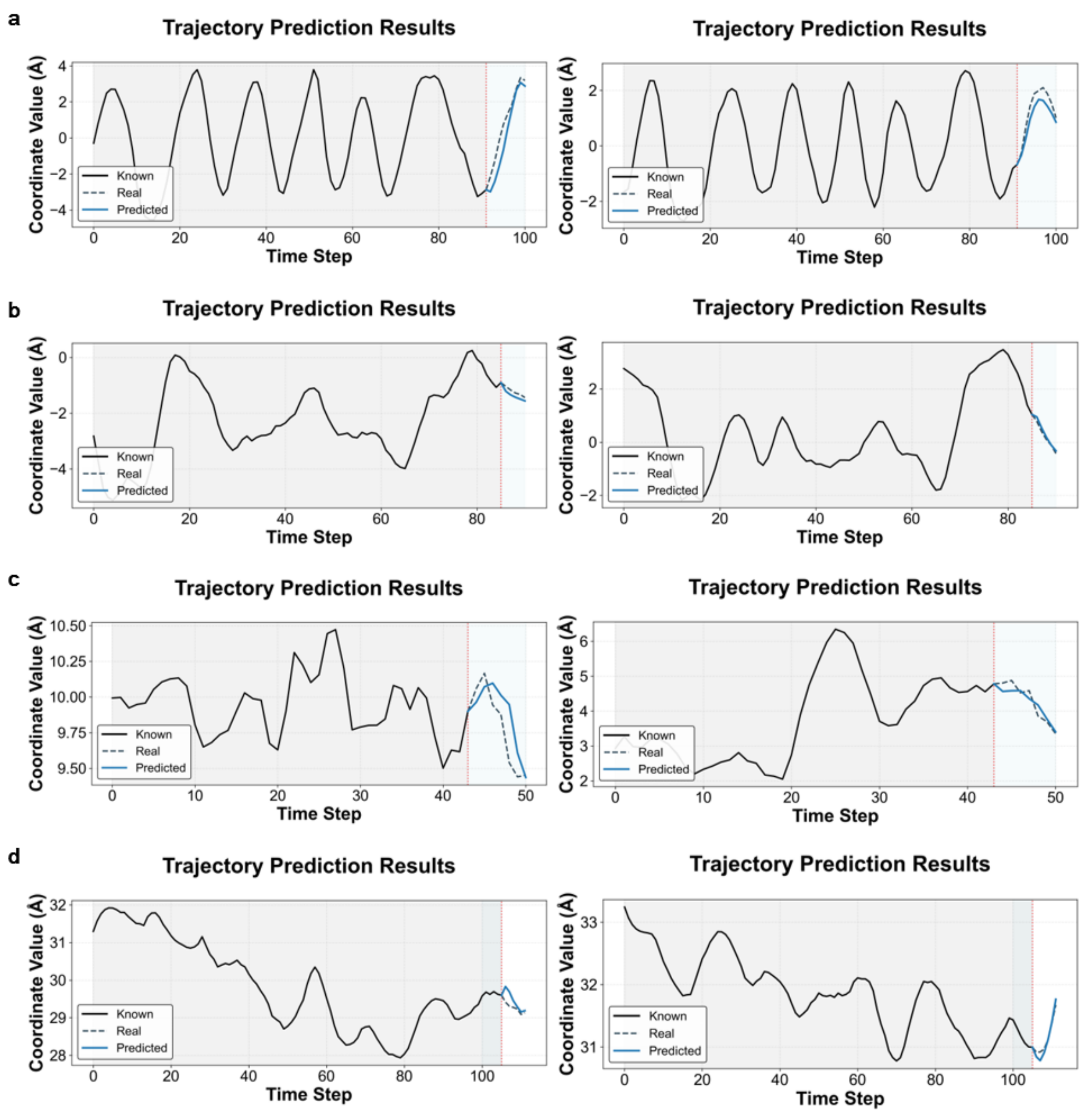

**Figure 3.** Prediction results of the model across four different biomolecular systems. a. Tetrapeptide. b. DHA. c. Stachyose. d.19-peptide.

Analysis of the tetrapeptide predictions (60 trajectories) revealed several key patterns. The model demonstrated strong overall performance in capturing trajectory variation trends, evidenced by a median Pearson correlation of 0.888. However, the model's performance varied significantly across different trajectories, suggesting that its generalization capability requires further improvement. NRMSE achieved high precision in most cases after normalization. However, it showed the largest standard deviation (0.664), and the mean value (1.132) substantially exceeded the median, revealing right-skewed distribution. A similar pattern of skewed distribution was observed for both Pearson and RMSE.

All evaluation metrics consistently demonstrated that, despite strong performance on most trajectories, the overall average was substantially reduced by a small subset of outliers with poor predictive accuracy. Complete per-trajectory results are available in Supplementary Information.

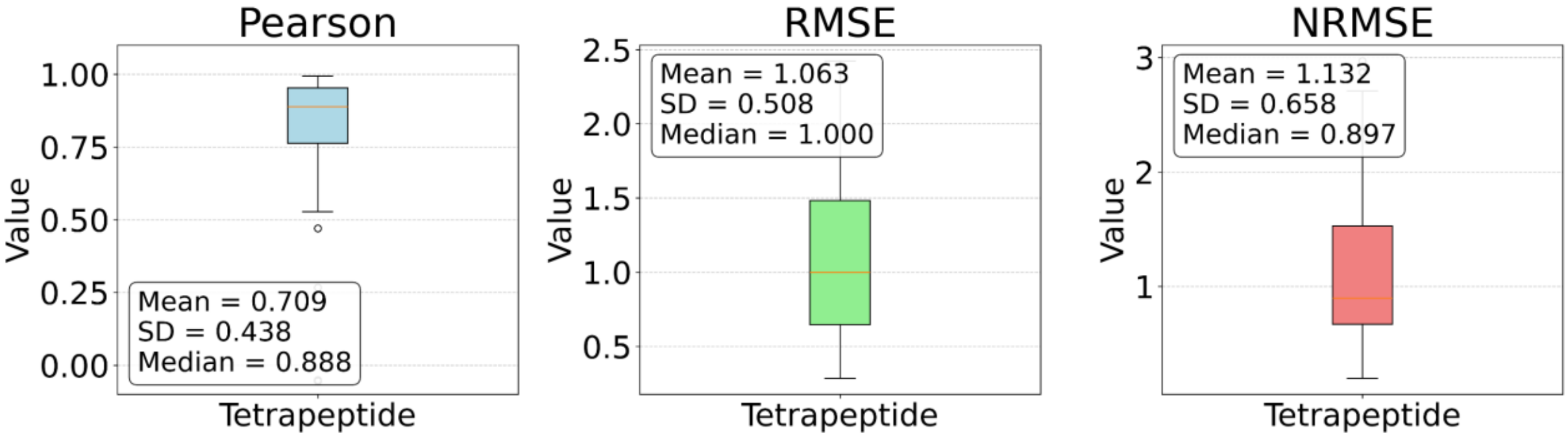

**Figure 4**. Distribution of model performance metrics. The box plots summarize the values of the Pearson correlation, RMSE, and NRMSE computed from 60 dimension. Descriptive statistics (mean, standard deviation, and median) for each metric are displayed within the corresponding plots.

By reconstructing each predicted trajectory into its corresponding three-dimensional molecular structure, we observed that the model generated reasonable molecular conformations. The proposed model dramatically reduced the computational cost while producing structures that exhibited high similarity to conformations generated by quantum-mechanical molecular dynamics simulations. The computation time was shortened from several days to just tens of minutes or a few hours.

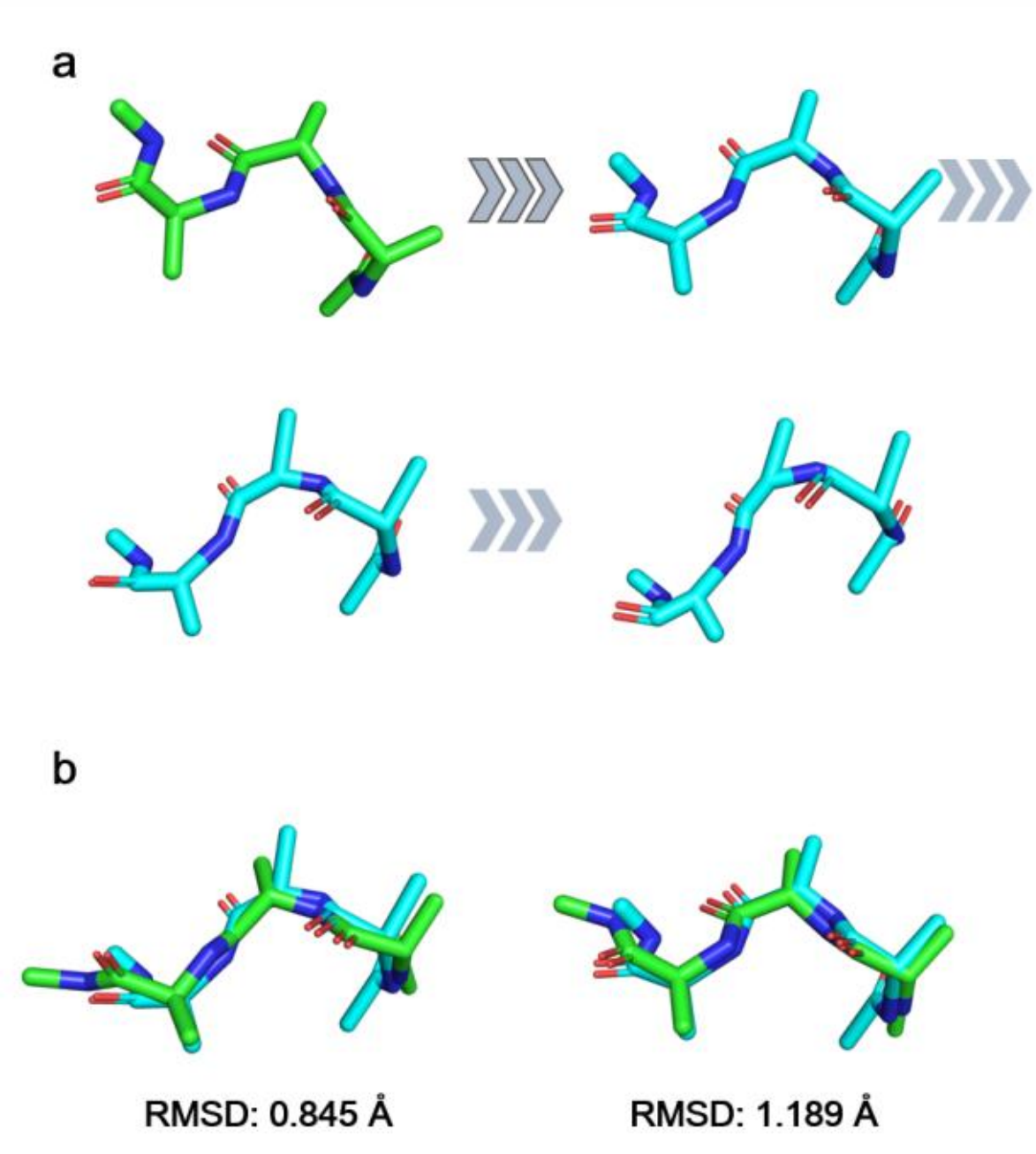

**Figure 5.** Predicted trajectory reconstruct into its corresponding three-dimensional molecular structure. a. Representative three-dimensional molecular structures reconstructed from the trajectory predicted by the ASTEROID model. b. Structural superposition of a predicted conformation (cyan) from ASTEROID and the corresponding reference structure (green) from quantum mechanics-based molecular dynamics (MD) simulation.

### Comparison results

As shown in Table 2, the proposed model consistently outperformed all eight baseline methods across all evaluation datasets. This demonstrated its strong applicability for molecular dynamics simulation data. All comparisons were conducted following identical training-validation-test splits to ensure fairness.

**Table 2.** Performance comparison with baseline models [a]

| Dataset | Metric | ASTEROID | HES | LSTM | RNN | SVR | STNN | STICM | DLINEAR | PatchTST |
|---|---|---|---|---|---|---|---|---|---|---|
| Tetrapeptide | Pearson | **0.706** | 0.008 | 0.527 | 0.456 | 0.494 | 0.529 | 0.341 | 0.553 | 0.371 |
| | RMSE | **1.063** | 2.003 | 1.953 | 1.571 | 1.570 | 1.669 | 1.815 | 1.642 | 2.330 |
| | NRMSE | **1.132** | 2.112 | 2.037 | 1.648 | 1.684 | 1.837 | 2.286 | 1.880 | 2.584 |
| Stachyose | Pearson | **0.361** | 0.010 | -0.207 | 0.145 | 0.015 | 0.174 | 0.037 | 0.261 | 0.117 |
| | RMSE | **0.694** | 0.755 | 0.773 | 0.823 | 0.849 | 0.736 | 0.895 | 0.771 | 0.776 |
| | NRMSE | 1.870 | 2.071 | 2.000 | 2.207 | 2.279 | **1.776** | 2.813 | 1.900 | 1.950 |
| 19-peptide | Pearson | **0.513** | **0.513** | 0.247 | -0.324 | -0.050 | 0.415 | 0.304 | -0.281 | -0.030 |
| | RMSE | 0.626 | **0.499** | 1.015 | 1.017 | 1.171 | 0.644 | 1.879 | 0.733 | 0.872 |
| | NRMSE | **3.672** | 5.454 | 8.017 | 7.717 | 10.168 | 3.710 | 13.701 | 4.154 | 10.525 |
| DHA | Pearson | **0.445** | 0.192 | 0.249 | 0.384 | 0.324 | 0.404 | -0.181 | 0.179 | 0.096 |
| | RMSE | **0.645** | 1.376 | 1.579 | 0.896 | 1.255 | 0.774 | 2.471 | 0.764 | 0.964 |
| | NRMSE | **2.906** | 6.547 | 7.785 | 3.896 | 5.640 | 3.676 | 12.804 | 3.891 | 4.848 |

[a] The best and the second-best results in each case are shown in bold and underlined, respectively.

### Iteration prediction

While the DEFM model achieves long-term forecasting by recursively applying a trained model with dynamically updated inputs[46], its iterative scheme relies on combining predicted target variables with true future observations of other variables to construct pseudo-observations. This approach becomes infeasible in practical settings such as molecular dynamics simulations, where access to actual future coordinates of all atoms is impossible during real prediction scenarios.

To overcome this fundamental limitation, we developed a fully autonomous iterative prediction strategy. In our approach, all variables are simultaneously forecast in the first round, and only these predicted values are used to construct subsequent input matrices—without relying on any ground-truth data. This self-contained mechanism advances multi-step prediction solely through the model' s own outputs, making it particularly suitable for modeling closed-system evolution.

After the first prediction round, the input for the second round ($X_{test}$) was updated using the predicted values from the first round, with no model retraining. We applied this method to the tetrapeptide data and successfully predicted the entire dataset, achieving a 72-step forecast based on an initial 91-step input. The performance of the tetrapeptide model in each iterative round is shown below.

**Table 3.** The performance of the tetrapeptide model in each iterative round.

| Iteration | Pearson *R* | RMSE | NRMSE |
|---|---|---|---|
| 1 | 0.710 | 1.063 | 1.132 |
| 2 | 0.658 | 1.256 | 1.514 |
| 3 | 0.715 | 1.210 | 1.619 |
| 4 | 0.711 | 1.186 | 1.648 |
| 5 | 0.716 | 1.165 | 1.504 |
| 6 | 0.720 | 1.234 | 1.488 |
| 7 | 0.748 | 1.247 | 1.457 |
| 8 | 0.569 | 1.124 | 2.470 |

We evaluated the integrated trajectory from 8 prediction rounds against ground-truth molecular dynamics simulations. As shown in the figure below, the model demonstrated robust performance across all trajectories, achieving a Pearson correlation of 0.762, RMSE of 1.280, and NRMSE of 0.806. These consistent metrics, combined with the trajectory visualizations presented, confirm the model's capability to accurately capture long-term molecular dynamics.

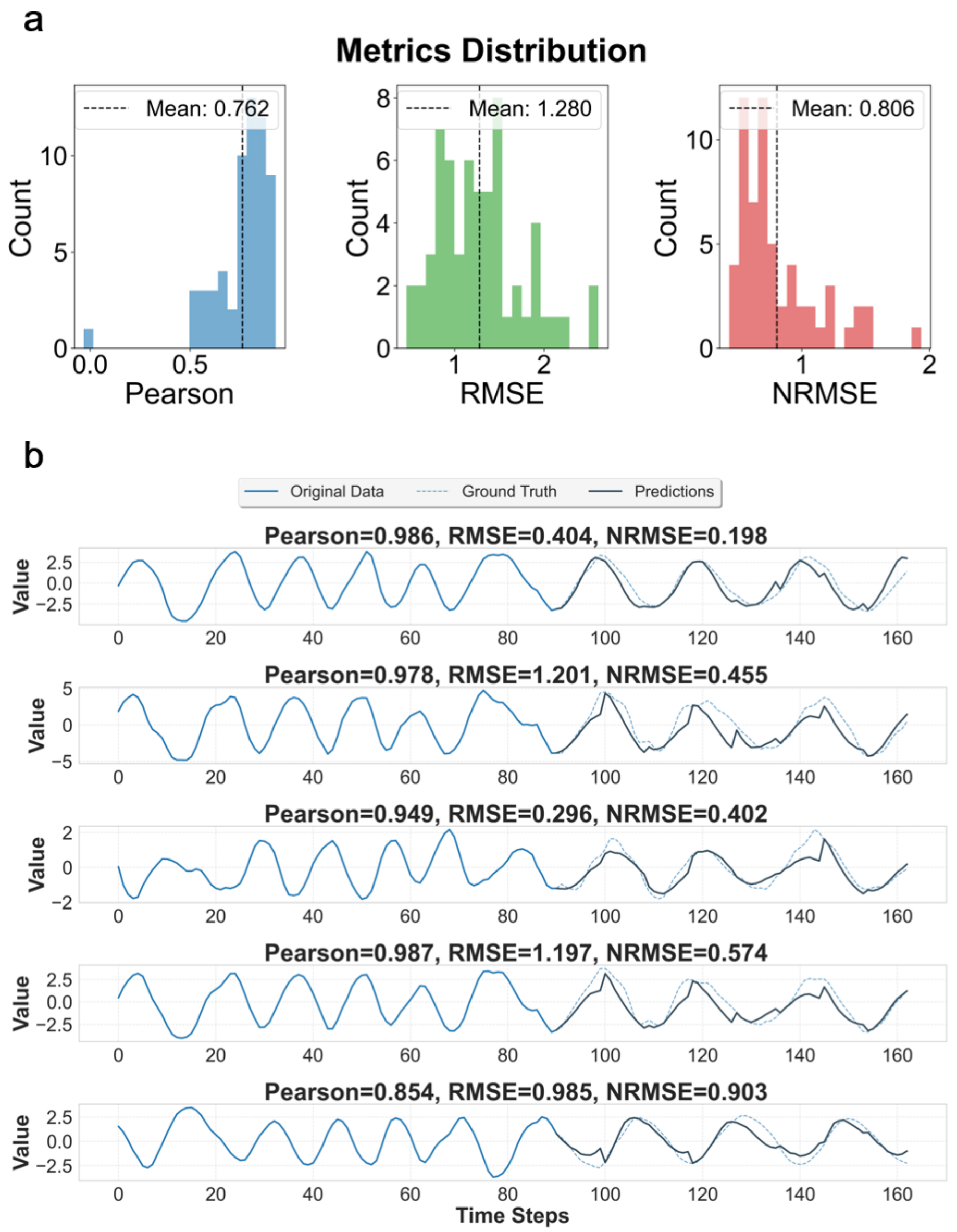

**Figure 6.** Quantitative and qualitative comparison between the ASTEROID model's predictions and the ground-truth molecular dynamics (MD) trajectory. a. Overall performance evaluation. The combined trajectory from 8 prediction rounds was compared against the full MD simulation. The distribution of evaluation metrics demonstrates the model's strong predictive capability. b. Visualization of partial trajectories. The solid blue line represents the input values. The black line denotes the predicted values, while the dashed blue line indicates the true values. Different background colors distinguish between iterative prediction rounds.

## 3.4 Result and analysis

### Robustness

To assess the model's robustness, we trained multiple models using data processed with varying Savitzky-Golay filter settings (different window lengths and polynomial orders). These settings produced trajectories with different noise levels and smoothness. Results showed that despite these differences in input data, all models performed consistently well on the same test set. This indicates that our approach was not sensitive to specific data preprocessing patterns, but instead learned robust feature representations essential to the forecasting task, demonstrating notable robustness against variations in input data characteristics.

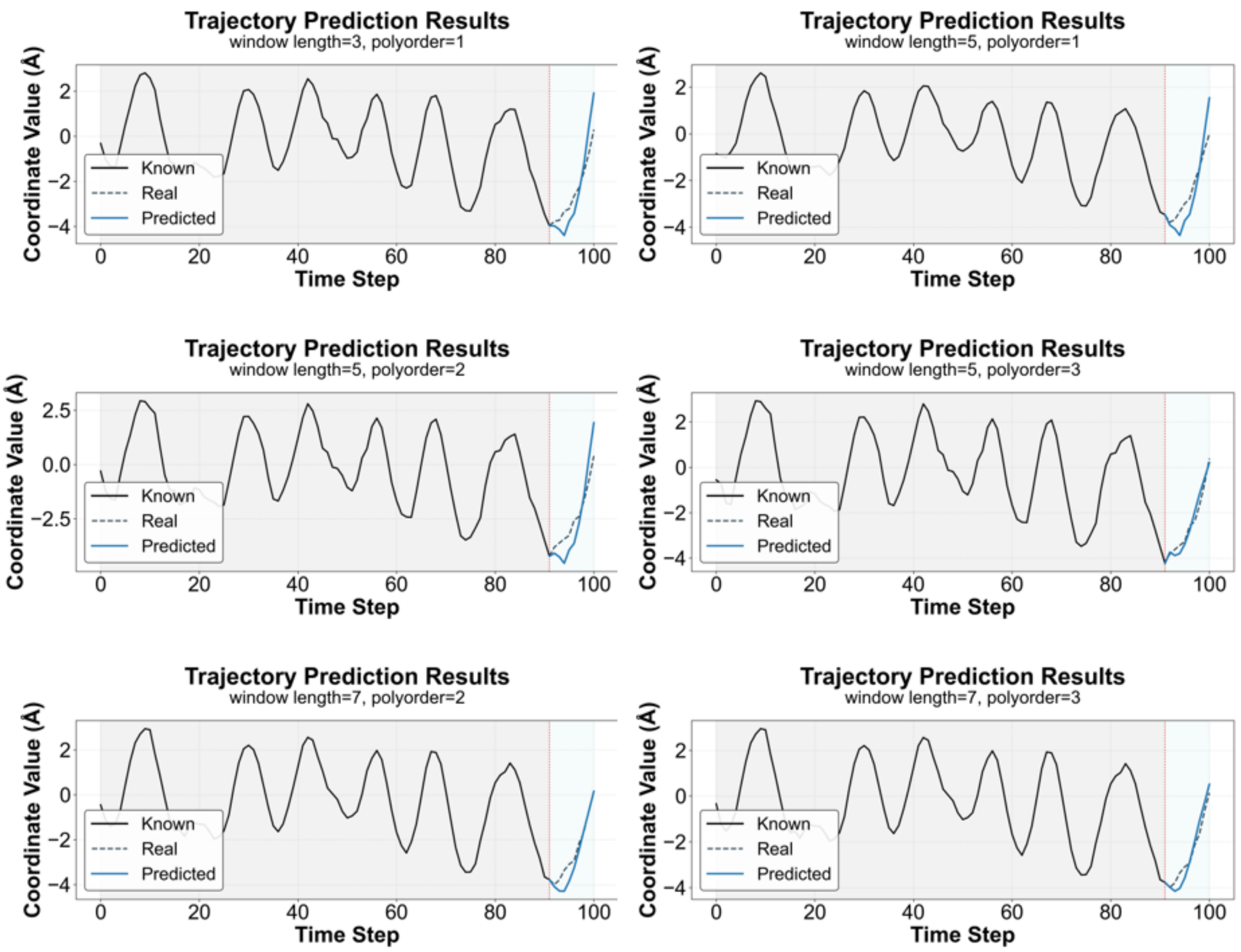

**Figure 7**. Model robustness under different data smoothing conditions. Despite being trained on data processed with varying Savitzky-Golay filter parameters, all models achieved consistent performance, demonstrating strong tolerance to preprocessing variations.

## Embedding dimension

According to Takens's theorem, when the embedding dimension L satisfies $L > 2d$ (where d is the intrinsic dimension of the dynamical system), the delay embedding could be topologically equivalent to the unknown phase space of the original system. To systematically evaluate the impact of embedding dimension on model performance, we first employed the box-counting method to estimate the intrinsic dimension. For the tetrapeptide dataset, this yielded an estimated dimension of 1.965. We constructed STI functions with different embedding dimensions L (ranging from 2 to $L_{max}$) and conducted systematic experiments to determine their optimal range. STI functions with varying embedding dimensions L were constructed, and the resulting low-dimensional time series data for different L values are illustrated in the corresponding figure.

$$\begin{bmatrix} y^1 & y^2 & \dots & y^m \\ y^2 & y^3 & \dots & \hat{y}^{m+1} \end{bmatrix} \dots \begin{bmatrix} y^1 & y^2 & & y^m \\ y^2 & y^3 & \vdots & \hat{y}^{m+1} \\ \dots & \dots & & \dots \\ y^L & \hat{y}^{L+1} & & \hat{y}^{m+L-1} \end{bmatrix}$$

In this experiment, the trends of prediction performance metrics, including the Pearson correlation coefficient, NRMSE, RMSE, with respect to the embedding dimension L reproduced the complete theoretical process from under-embedding to successful embedding and finally to over-embedding, providing strong empirical support for the theorem.

The evolution exhibited three distinct phases. When L failed to meet the theoretical requirement $L>2d$, phase space reconstruction was incomplete. Consequently, the model could not extract deterministic dynamical rules from the input vectors, resulting in severely degraded forecasting accuracy. This was consistent with the experimental results, which showed an extremely low Pearson coefficient (–0.067) and a high NRMSE (9.268), indicating complete predictive failure. When L reached the optimal range satisfying $L>2d$, the model successfully captured the system's underlying dynamics, achieving peak performance with a maximum Pearson correlation (0.743) and minimum NRMSE (0.985). Further increasing L beyond this range, while theoretically valid, led to performance degradation due to the expanding prediction horizon and reduced effective training data, evidenced by a decline in Pearson correlation and a rebound in NRMSE.

Within the optimal range, RMSE remained stable or decreased slightly, indicating that the accurately reconstructed dynamics provided sufficient information to mitigate error accumulation during iterative prediction. This confirms that the model performs high-quality forecasting in a properly reconstructed phase space rather than simple numerical extrapolation.

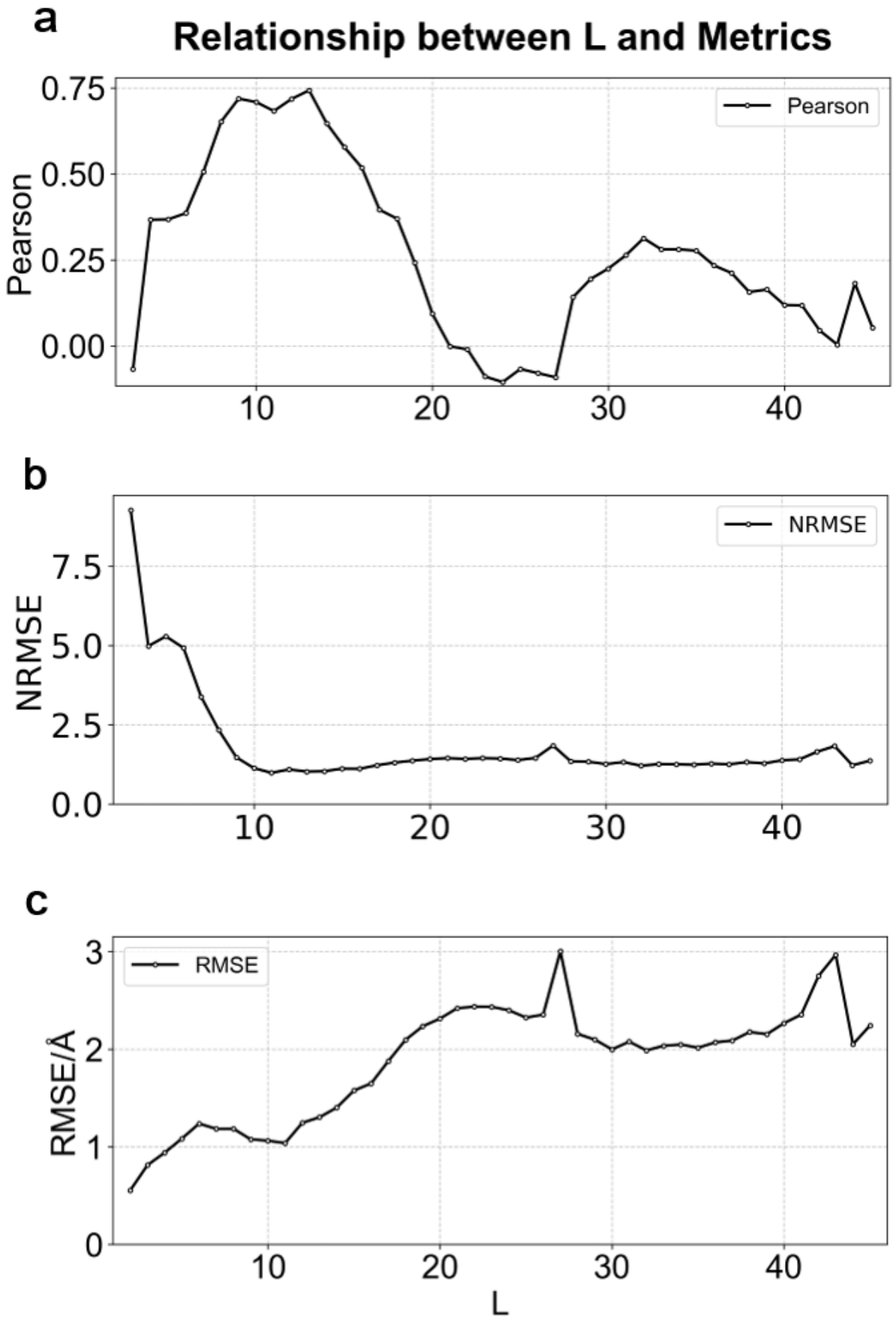

**Figure 8.** Evolution of prediction performance with embedding dimension L. The coordinated trends of (a) Pearson coefficient, (b) NRMSE, and (c) RMSE collectively reproduce the complete theoretical trajectory from under-embedding through successful embedding to over-embedding, empirically validating the critical role of embedding dimension in dynamical system reconstruction.

To quantitatively evaluate the systematic impact of embedding dimension on the prediction performance of ASTEROID model, we conducted a statistical analysis of NRMSE across three embedding dimension regions under different observation time steps (total time steps ranging from 50 to 100). Prediction error in Area II was markedly lower and exhibited the smallest variance, indicating optimal and highly stable performance in this region.

We further applied the Kruskal−Wallis test to verify the statistical significance of inter-region differences, obtaining an H-statistic of 14.377 with a highly significant p-value of $7.551\times10^{-4}$. This p-value proved that the NRMSE distributions across the three regions were highly significantly different.

This statistical conclusion strongly supported the aforementioned theoretical analysis. The embedding dimension had a significant dual impact on prediction performance. In Area I, NRMSE was high, as L is below or close to the minimum dimension required by Takens's theorem, resulting in a reconstructed phase space that could not fully preserve the topological information of the original system. The model failed to capture the complete system dynamics, leading to poor prediction performance. Area II provided the optimal L range, where complete phase-space reconstruction enabled full historical information usage and minimal NRMSE. In Area III, despite theoretical sufficiency, prediction accuracy declined as the high dimension induced a data sparsity problem and increased model complexity, resulting in overfitting and heightened noise sensitivity.

Overall, the initial decline and subsequent rise of NRMSE reflect a consistent and non-random pattern, confirming that the model delivers stable and accurate predictions only within a well-defined embedding range. Moreover, the extent of this suitable range increases with the volume of observational data.

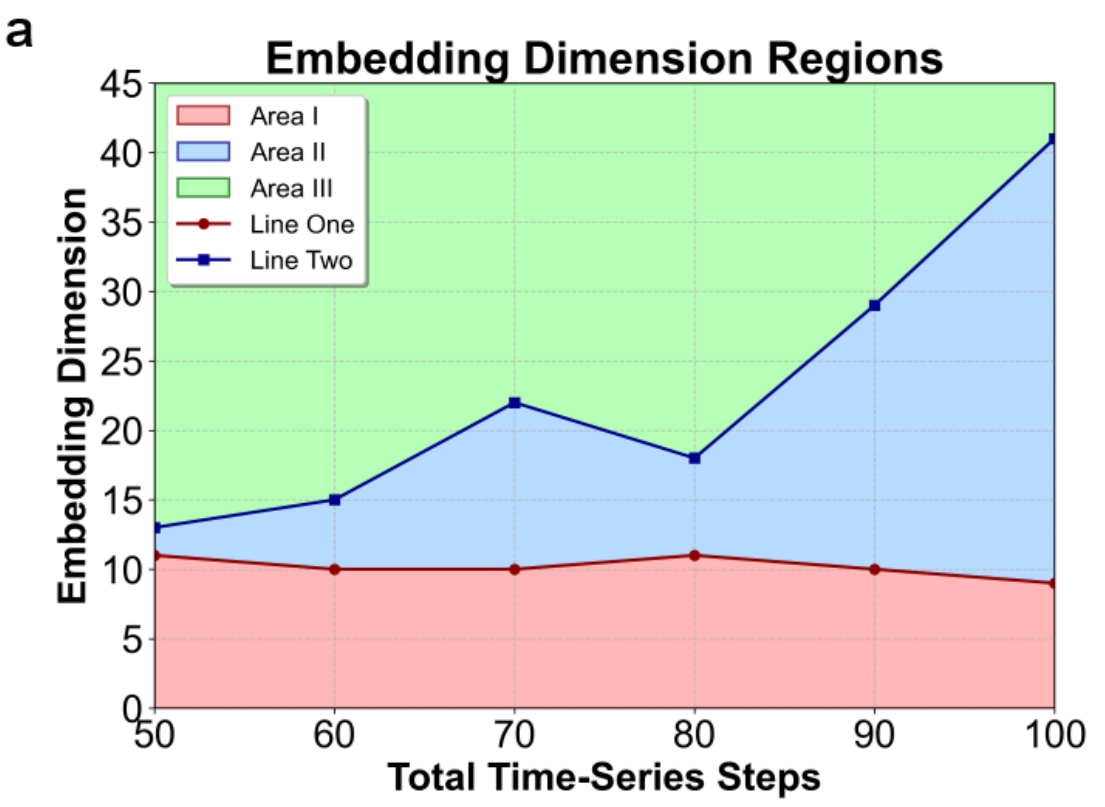

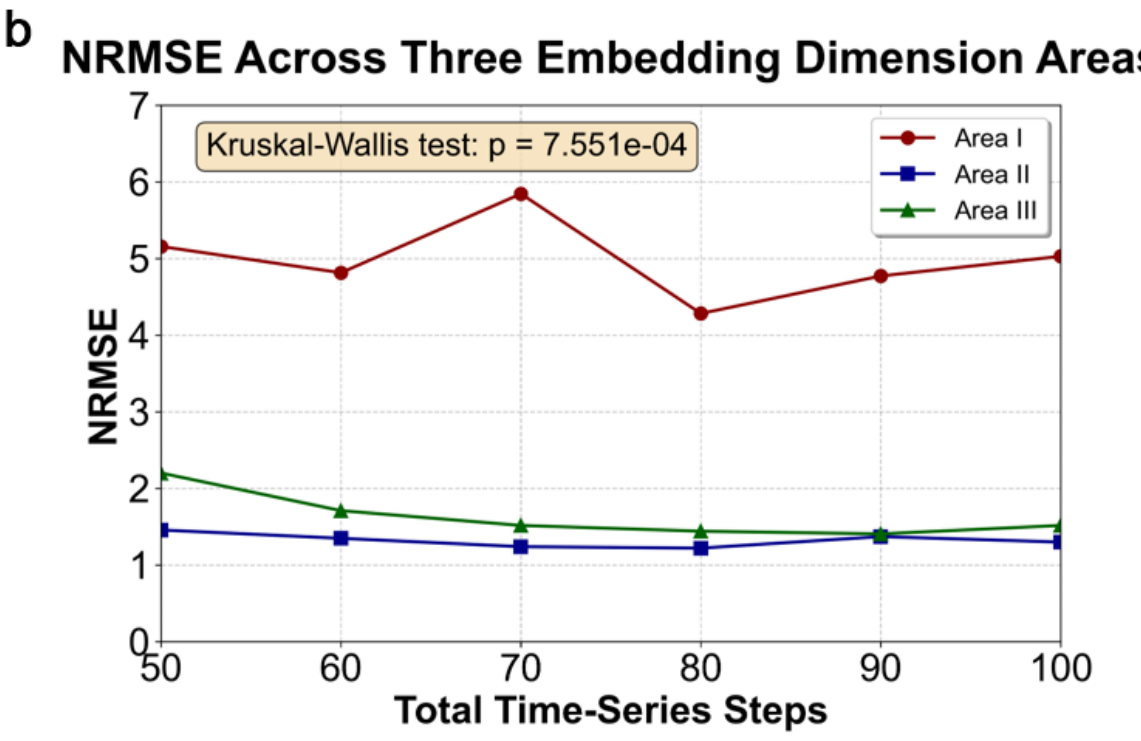

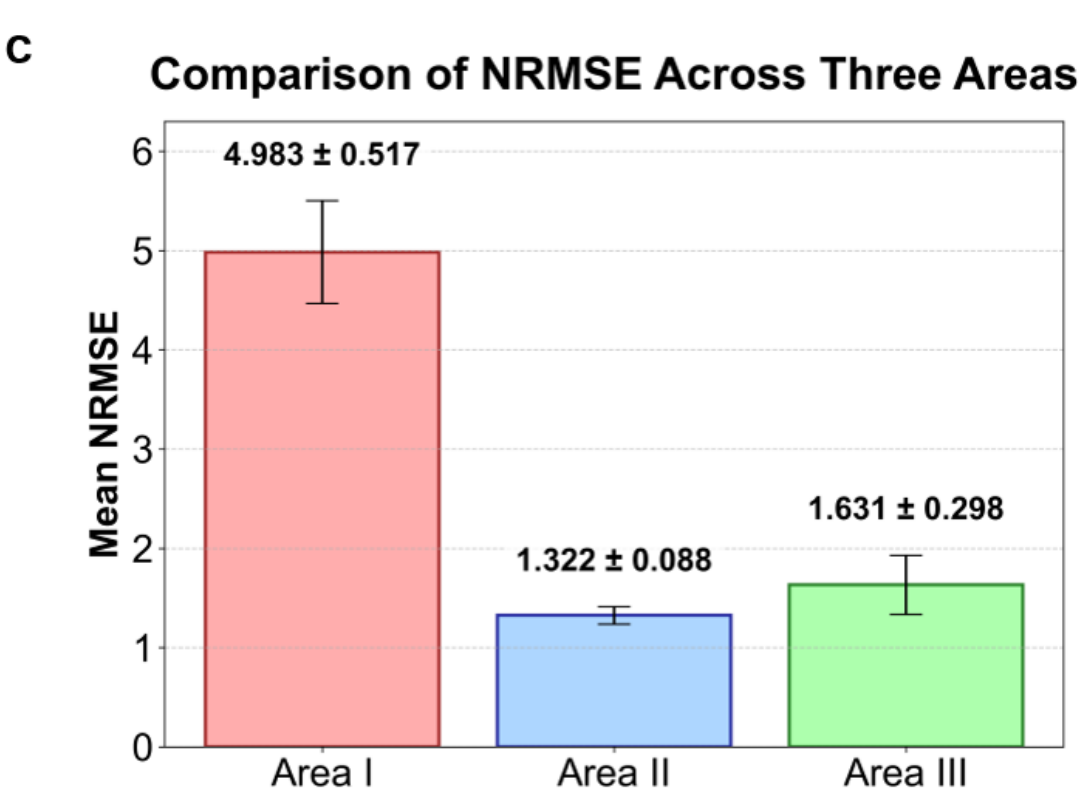

**Figure 9.** Identification of the optimal embedding dimension region and its statistical validation across varying observation lengths. a. Definition of embedding dimension regions. Area I (Under-embedding), Area II (Optimal), and Area III (Over-embedding) are delineated based on the total observable time steps. b. NRMSE trends across regions and observation lengths. c. Prediction error (NRMSE) is consistently lowest and most stable in Area II compared to Areas I and III, demonstrating the universal advantage of optimal embedding across different data volumes. c. Statistical significance of regional performance differences.

## Ablation experiment

To evaluate the contribution of each component, we conducted systematic ablation studies on the three attention mechanisms: temporal self-attention, short-range spatial attention, and long-range spatial attention. The experimental protocol sequentially removed each mechanism individually, then in pairs, and finally replaced the entire proposed attention block with standard multi-head attention. All models were evaluated on the tetrapeptide dataset, with results averaged over all trajectories, and were trained under identical settings to ensure a fair comparison.

The ablation results are summarized in Fig. 10 using NRMSE. The complete ASTEROID model achieved the lowest NRMSE, validating the effectiveness of its integrated three-component attention design. Further analysis revealed that models retaining only spatial attention outperformed those with only temporal attention, indicating that atomic motion prediction depends more critically on spatial context than on temporal dynamics. This finding aligns with the physical principle that atomic interactions are governed primarily by spatial proximity. Replacing the specialized attention mechanisms with standard multi-head attention caused severe performance degradation (NRMSE = 17.582), unequivocally demonstrating the necessity of our task-specific architectural design.

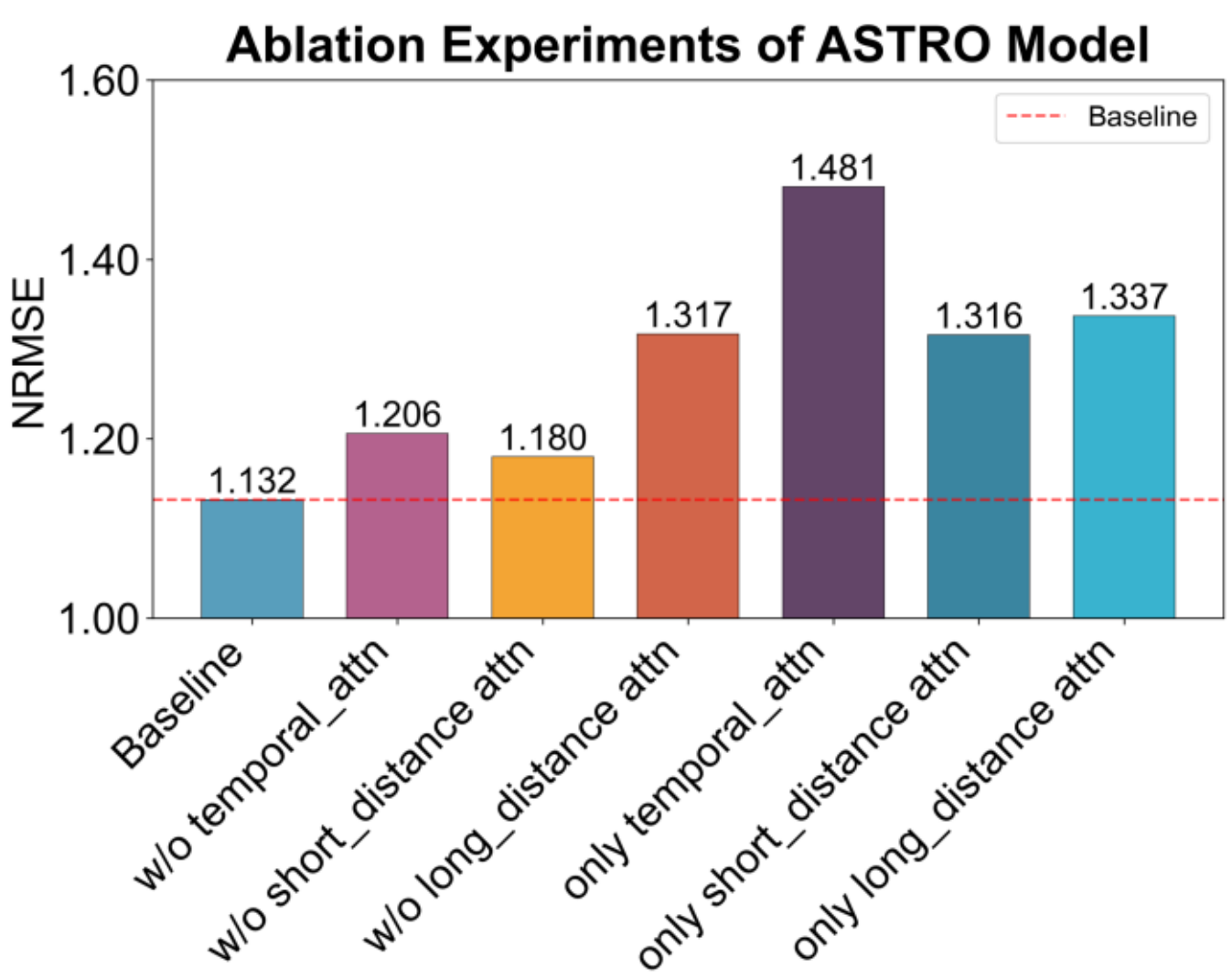

**Figure 10.** Ablation study on the spatiotemporal attention mechanisms. The contribution of each attention component (temporal, short-range spatial, long-range spatial) is evaluated by systematically removing or replacing them. The complete ASTEROID model achieves the best performance.

## 5. CONCLUSION

This study addresses the fundamental challenge in molecular dynamics of balancing computational efficiency with simulation fidelity. We present a novel approach that models simulation trajectories as high-dimensional spatiotemporal sequences, leveraging time-series forecasting to enable efficient, long-term prediction. To capture the inherent nonlinearity and high dimensionality of molecular systems, we introduce the Spatiotemporal Information (STI) Transformation equation and integrate it into a Transformer-based encoder-decoder architecture, forming the ASTEROID framework. This framework has been validated on four representative biomolecular systems, demonstrating robust predictive capabilities and generalization performance.

The model integrates both local and global attention mechanisms in a spatial context. This dual approach enables the capture of short-range physical interactions as well as long-range functional correlations between atoms, thereby facilitating a comprehensive understanding of complex structural features. In terms of temporal dynamics, it combines the encoder's global context modeling with the decoder's autoregressive causal constraints. The iterative autoregressive prediction framework, along with dynamic sliding window inference, ensures adherence to causality while enabling stable long-horizon forecasting. Furthermore, we introduce an embedding method that effectively incorporates positional and semantic information.

Experimental results demonstrate that ASTEROID significantly reduces the computational cost of molecular dynamics simulations while maintaining high trajectory accuracy over extended horizons, offering a scalable tool for studying biomolecular dynamics. Its novel iterative inference scheme enables long-term predictions without retraining. The model demonstrates strong robustness across varying noise levels and preprocessing parameters. Additionally, by systematically varying the embedding dimension, we successfully reproduce the entire theoretical process from under-embedding to successful embedding and subsequently to over-embedding. Statistical analysis confirms significant performance differences across these regimes, demonstrating a clear and interpretable link between embedding dimension and model performance.

While ASTEROID demonstrates robust predictive performance in most test cases, its variability across different trajectories currently limits practical reliability. This inconsistency likely arises from: (1) a training data distribution that insufficiently captures the full spectrum of molecular motion patterns; (2) inherent architectural constraints in modeling certain complex dynamical features. Furthermore, as ASTEROID's theoretical framework is primarily based on deterministic dynamics, it cannot make accurate predictions for strongly noisy data.

Future work will focus on enhancing model stability and accuracy, particularly for challenging trajectories where performance remains suboptimal. In light of practical applications, subsequent research will concentrate on further incorporating physical priors, extending investigations into larger system simulations, and exploring efficient conformation space sampling through generative architectures to advance computational molecular simulation methodologies.